\documentclass{amia}
\usepackage{lipsum} 
\usepackage{graphicx}
\usepackage{booktabs}
\usepackage{xcolor}
\usepackage{hyperref}
\hypersetup{hidelinks}
\begin{document}

\title{Grounded Adjudication of Variations across Extracted TimeLines (GAVEL): Comparing Clinical Timelines Against Their Case Reports}

\author{Jack Cummins$^{1,*}$, Sayantan Kumar, PhD$^{2,*}$, Ketan Tamirisa$^3$, Jeremy C. Weiss, MD, PhD$^2$ }

\institutes{
    $^1$ Princeton University, Princeton, New Jersey, USA \\
    $^2$ National Library of Medicine, Bethesda, Maryland, USA \\
    $^3$ Washington University in St. Louis, St. Louis, Missouri, USA\\
    * denotes equal contribution
}

\maketitle

\section*{Abstract}
\vspace{-2.5mm}
Existing pipelines for clinical timeline extraction from case reports are evaluated using an expert reference and are limited by imperfect reference annotations and imprecise event alignment. We developed GAVEL, an LLM-judge protocol that compares two timelines with the case report and returns a discrepancy type, verdict, and report passage for each difference. We evaluated the event matcher, reviewed 2,738 findings from GPT-5.6-sol and DeepSeek V3.2, ranked six LLM extractors and two human annotators, and tested GAVEL-guided merging. True-match rates were 60\% immediately below and 48\% immediately above the 0.10 cutoff. Manual review confirmed 89.4\% and 88.6\% of findings. Across 126 reports, merged timelines were preferred in 77.0\% of comparisons (95\% CI, 69.8–84.1\%) and reduced discrepancies attributed to the evaluated timeline from 7.63 to 0.85 per report. GAVEL supports report-based comparison and revision without treating either timeline as ground truth.

\section*{Introduction}
\vspace{-3mm}

Clinical narratives encode both the events in a patient’s course and their timing. Incorrect temporal placement can distort a disease trajectory, reverse the apparent order of events, or bias time-to-event estimates. Accurate temporal representations are important for computational phenotyping and longitudinal modeling \cite{estiri2021temporal}. Recent improvements in large language models (LLMs) have made it possible to extract structured relative timelines directly from full-length case reports, scaling from expert-reviewed studies to 124,699 reports and more than 5.6 million timestamped events \cite{wang2025large,noroozizadeh2025pmoatts}.
To evaluate these timelines, prior studies compared extracted timelines with clinician-authored references.
Candidate and reference events are aligned using semantic similarity, after which event recovery and temporal agreement are summarized with standardized metrics \cite{wang2025large,noroozizadeh2025pmoatts,kumar2026glp1ra}.
This framework provides a reproducible basis for comparing extractors, but it treats the selected annotation as correct for scoring.

Human annotation remains essential for clinical evaluation, yet an independently constructed timeline is not necessarily an error-free representation of the source. In our study, two clinically trained annotators independently constructed timelines from the same case set of reports, and the measured performance of LLM extractors differed depending on which annotator served as the reference \cite{kumar2026glp1ra}.
When a model and an annotator disagree, these metrics cannot determine which timeline is better supported by the report, whether both interpretations are supportable, or whether the report itself is
ambiguous. One timeline must therefore be designated as ground truth before scoring begins.

Another source of error arises when candidate and reference events are aligned.
In prior work, candidate and reference events are recursively matched using cosine distance under a fixed threshold \cite{wang2025large,kumar2026glp1ra}.
Our audit shows substantial interleaving of true and false event matches near the operational cutoff.
Consequently, a final timeline score may reflect two unresolved decisions made before temporal correctness is assessed: whether the chosen annotation should be treated as ground truth and whether the candidate and reference events were aligned correctly.
Aggregate metrics summarize agreement after alignment, but they do not identify why two timelines differ---for example, because of an omitted event, an unsupported addition, an incorrect value or time, an annotation error, or insufficient evidence in the report.

An evaluation method for this setting should address both limitations described above: dependence on a selected reference timeline and errors introduced when candidate and reference events are aligned. To do so, it would need to compare two timelines without selecting either as the reference, use the original case report to match corresponding events, adjudicate differences in event inclusion, values, and timing, and retain the report text supporting each decision.
Existing LLM evaluation methods provide these capabilities separately. Pairwise LLM judges compare and rank two outputs \cite{zheng2023judging,chiang2024chatbotarena}; factuality methods verify generated claims against a knowledge source or patient record \cite{min2023factscore,chung2025verifying}; and TIMER evaluates temporal reasoning over longitudinal clinical records \cite{cui2025timer}. To the best of our knowledge, no prior method combines these capabilities to adjudicate event-level differences between two complete extracted clinical timelines.

We introduce \textbf{GAVEL} (\textbf{G}rounded \textbf{A}djudication of \textbf{V}ariations across \textbf{E}xtracted time\textbf{L}ines), an LLM-based framework that compares two extracted timelines using their originating case report.
Given a case report and two candidate timelines, A and B, GAVEL links each event to report text, matches corresponding
events across the timelines, compares values and relative times, and classifies each remaining discrepancy.
Every finding contains a discrepancy type, a verdict indicating whether the report supports A, B, both, neither, or cannot resolve the difference, and the report passage used for the verdict or an explicit indication that no supporting passage was found. This lets a reviewer inspect the evidence for each verdict. The case report determines the verdict, and
neither candidate timeline serves as the reference. The same procedure can therefore compare human- and model-generated timelines and rank them on a common scale.

\textbf{Contributions.}
\textbf{First}, we audit the PubMedBERT event-matching procedure underlying existing timeline metrics using 40,259 aligned event pairs and
manually review a stratified sample of 100 pairs against their case reports.
\textbf{Second}, we define the GAVEL protocol using discrepancy types and verdicts accompanied by the report text used to support them.
\textbf{Third}, we assess two judge models through manual review of 2,738 findings, automated checks that quoted passages were present in the corresponding case reports, and independent review of a 150-finding subset.
\textbf{Fourth}, we use GAVEL to compare eight timeline sources: six LLM extractors and two human annotators, followed by a corpus-scale comparison of
the highest-rated human annotator, proprietary model, and open-weight model.
\textbf{Finally}, we use findings supported by the case report to revise the timeline produced by the highest-scoring model (GPT5.6-sol), and evaluate the revised timeline with a different judge model.

\section*{Related Work}
\vspace{-3mm}
We compare prior methods along three capabilities relevant to GAVEL: pairwise comparison, use of the originating clinical text, and explicit

\textbf{LLM judges and pairwise evaluation.}
LLM-based evaluators provide an alternative to lexical and reference-based metrics for open-ended generation. G-Eval demonstrated structured evaluation using explicit criteria \cite{liu2023geval}, while MT-Bench and Chatbot Arena established scalable
pairwise judging and Bradley--Terry-style model comparison \cite{zheng2023judging,chiang2024chatbotarena}.
These judges can be sensitive to candidate position and response verbosity, and may favor outputs produced by their own model family \cite{zheng2023judging,panickssery2024selfpreference}. These biases motivate direct validation of the judge.
These methods make large-scale pairwise evaluation feasible, but they usually assess the overall quality of two responses rather than whether individual clinical events are supported by the source document. GAVEL uses pairwise judging to resolve specific event, value, and timing differences between two timelines using the originating case report.

\textbf{Factuality and clinical evaluation.}
Factuality methods evaluate generated content against a specified evidence source. FActScore decomposes generated text into atomic facts and verifies them against a knowledge source \cite{min2023factscore};
VeriFact similarly retrieves patient-specific EHR evidence to classify propositions in generated clinical documents \cite{chung2025verifying}.
These approaches evaluate one generated document at a time. TIMER addresses temporal reasoning over longitudinal EHRs and validates LLM-generated correctness and completeness scores against clinician rankings, but evaluates free-text answers rather than event-level disagreements between two timelines \cite{cui2025timer}. 
Pairwise LLM judges typically compare two outputs at the response level, whereas FActScore and VeriFact assess claims in a single generated output and TIMER scores free-text temporal responses. GAVEL uses a different unit of analysis. It matches events across two structured timelines and returns a separate verdict, together with the supporting case-report
passage, for each difference in event inclusion, value, or timing. To the best of our knowledge, prior work has not evaluated two complete extracted clinical timelines in this form without designating either timeline as
the reference.



\section*{Methods}
\vspace{-1mm}
\subsection*{Study Corpus and Timeline Representation}
\vspace{-3mm}
We used the glucagon-like peptide-1 receptor agonist (GLP-1RA) case-report corpus from our prior work \cite{kumar2026glp1ra}, which provides independently constructed timelines from two clinically trained annotators and multiple LLM extractors. Each timeline is a sequence of $(e_i,t_i)$ pairs, where $e_i$ is a free-text clinical event and $t_i$ is its time in hours relative to an index time $t=0$; negative and positive values denote events before and after the index encounter, respectively. 

We followed the published temporal conventions \cite{wang2025large,kumar2026glp1ra}: $t=0$ was the first documented presentation, admission, or other index encounter, or the principal diagnosis or treatment when no encounter was stated; intervals were assigned to their onset; and histories without a derivable onset were placed at the encounter. These conventions were applied identically to both LLM-generated and human-annotated timelines. 
We excluded six reports in which at least one event required for the evaluated clinical course appeared only in a figure that was unavailable to the annotators.
This yielded 126 reports for the corpus-scale analyses. Source texts were obtained from plain-text files in the December 17, 2024 PubMed Open Access \texttt{oa\_noncomm} release.

\vspace{-2mm}
\subsection*{Audit of Reference-Based Event Matching}
\vspace{-2mm}
Prior timeline evaluations recursively align candidate and reference events using PubMedBERT cosine distance and retain alignments below 0.10 \cite{gu2021domain,wang2025large,kumar2026glp1ra}. We pooled the candidate--reference event pairs assigned a finite cosine distance by this matching procedure across the corpus and removed repeated event-text pairs, yielding 40,259 unique pairs. We sampled 25 pairs from each of four prespecified bands, $[0,0.08)$, $[0.08,0.10)$, $[0.10,0.12)$, and $[0.12,0.30)$, for 100 pairs total. A reviewer examined each pair in the context of the case report from which both timelines were extracted and labeled whether both descriptions referred to the same clinical event. We report band-specific true-match proportions with Wilson 95\% confidence intervals. 

\vspace{-2mm}
\subsection*{The GAVEL Adjudication Protocol}
\begin{figure}[t] 
\vspace{-1.5mm}
\centering 
\includegraphics[width=0.85\linewidth]{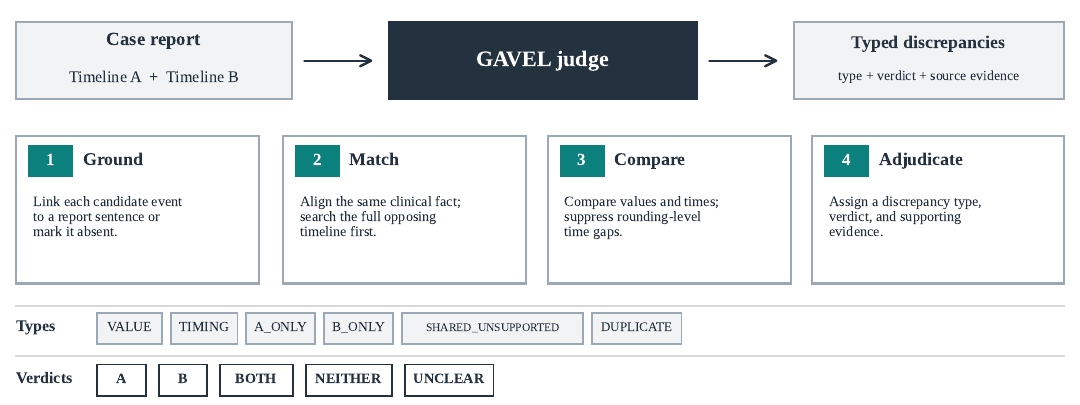} 
\caption{The GAVEL adjudication protocol. The LLM judge uses the case report to compare two candidate timelines: it links events to report text, matches corresponding events, and compares values and timestamps. For each discrepancy, it returns a type, a verdict, and the report sentence cited for that verdict, or indicates that no supporting sentence was found. Neither timeline serves as the reference.}
\label{fig:gavel_protocol} 
\end{figure}

\vspace{-2mm}
GAVEL receives the complete case report and two candidate timelines, labeled A and B (Figure~\ref{fig:gavel_protocol}). The prompt does not disclose how either timeline was produced, and neither is designated as the reference. The final prompt was held fixed across all reported cases, pairings, and judge models; no case-specific instructions or examples were added. GAVEL performs four dependent operations (Table~\ref{tab:worked_example}) and identifies a discrepancy after matching events across the two timelines. A \emph{discrepancy} is one difference in event inclusion, value, timestamp, source support identified after events in A and B are matched. A \emph{finding} is the structured JSON
object returned for one discrepancy.

\begin{table}[!htbp]
\vspace{2mm}
\centering
\caption{A worked GAVEL adjudication from report PMC6143713. Both timelines record the same event string, so the discrepancy is purely temporal. The report states the interval explicitly, and the judge returns a typed verdict linked to the source text supporting the verdict.}
\vspace{-2mm}
\label{tab:worked_example}
\footnotesize
\begin{tabular}{@{}p{0.2\linewidth}p{0.74\linewidth}@{}}
\toprule
\textbf{Source sentence} & ``During a telephone call 2 weeks after his visit, the patient described worsening erythema surrounding the biopsy site with yellow, thickened drainage. Thus, he was started on doxycycline, 100 mg twice daily for a week.'' \\
\addlinespace[3pt]
\textbf{Timeline A (GPT-5)} & \texttt{doxycycline 100 mg twice daily started} at $t=336$ h \\
\textbf{Timeline B (o4-mini)} & \texttt{doxycycline 100 mg twice daily started} at $t=504$ h \\
\addlinespace[3pt]
\textbf{Type} & \texttt{TIMING} \\
\textbf{Verdict} & \texttt{A}. The initial visit anchors $t=0$ and the report places the call two weeks later, giving 336 h. Timeline B's placement is three weeks after the visit and is unsupported. \\
\bottomrule
\end{tabular}
\vspace{-3mm}
\end{table}
 \textbf{1. Anchor (Ground)} For each event in A and B, the judge identifies the supporting sentence in the report that states the event or marks it as having no supporting sentence. This ground anchor is distinct from the temporal origin $t=0$ and is quoted only when the event contributes to a finding.

\textbf{2. Match.} Events in timelines A and B are matched when they describe the same clinical event and are linked to the same report sentence, even if their wording, value, or timestamp differs. Sharing a sentence alone is insufficient when that sentence describes multiple events; each event is matched separately. If neither event is found in the report, they are matched only when they make the same clinical assertion. Before
labeling an event one-sided, the judge searches the full opposing timeline. When one timeline records an event in one row and the other splits the same
event across several rows with the same source sentence and timestamp, those rows are treated as one match.

\textbf{3. Compare.} For each matched event pair the judge compares values and times across A and B, both of which are visible without the report. Times are treated as identical when they differ by less than the maximum of 24 hours or 10\% of the event's distance from t = 0. The 24-hour minimum reflects the day-scale temporal resolution used in this study, for which sub-day differences were not treated as distinct. The proportional term similarly treats small
relative differences for events farther from $t=0$ as equivalent.

\textbf{4. Classify and adjudicate.} Each finding receives exactly one of six discrepancy types: \texttt{VALUE}, for conflicting numeric values or other clinical attributes in a matched event;
\texttt{TIMING}, for timestamps that differ beyond the specified tolerance; \texttt{A\_ONLY} or \texttt{B\_ONLY}, for an event present in only one
candidate timeline; \texttt{SHARED\_UNSUPPORTED}, for an event present in both timelines but unsupported by the report; and \texttt{DUPLICATE}, for different numbers of occurrences across the two timelines.
For \texttt{A\_ONLY} and \texttt{B\_ONLY}, the type identifies which timeline contains the event; the verdict determines whether the event is supported by the report or is an unsupported addition.
A matched event that differs in both value and time produces separate \texttt{VALUE} and \texttt{TIMING} findings. 

Each finding receives exactly one of five verdicts:$\{\texttt{A},\texttt{B},\texttt{BOTH},\texttt{NEITHER},\texttt{UNCLEAR}\}$. \texttt{A} and \texttt{B} indicate that the report supports the corresponding timeline over the other. \texttt{BOTH} is restricted to cases in which both representations are consistent with the report because they differ only in wording or express the same value in different forms. \texttt{NEITHER} indicates that neither representation is supported, including a shared event that is absent from the report or whose content the report contradicts on both sides. \texttt{UNCLEAR} is used when the report remains insufficient to resolve the difference after applying the temporal conventions, including cases with an underivable event time or a genuinely indeterminate $t=0$. A timing disagreement alone is not sufficient reason to assign \texttt{UNCLEAR}.
The judge returns one JSON object per discrepancy. Each object contains the discrepancy type; the event text and time from each candidate, with null values for an absent side; the event polarity as present (finding occurred) or absent (finding did not occur or was not detected.); the verdict; a short rationale; and either a verbatim supporting span from the report or the literal value \texttt{none found}. 

\textbf{Judge models and inference.} We ran the same GAVEL prompt with two different judge models. GPT-5.6-sol was accessed through Azure OpenAI (API version \texttt{2024-12-01-preview}) with high reasoning effort and a 64,000-token output limit. DeepSeek V3.2 was served locally from the \texttt{DeepSeek-V3.2-UD-IQ2\_M} GGUF checkpoint using \texttt{llama.cpp}, greedy decoding, a 131,072-token context window, and a 100,000-token output limit. Instructions, report, and both timelines were submitted as one user message; model identities were omitted from the prompt.

\vspace{-2mm}
\subsection*{Four-Stage Evaluation Design} 
\vspace{-2mm}

\textbf{Stage 1: judge validation.} We evaluated 10 reports across 3 extractor pairings drawn from GPT-5, o4-mini, and Llama 3.3 70B, yielding 30 games for each judge (GPT-5.6-sol and DeepSeek V3.2).
For every finding, the primary reviewer examined the report and both timelines,recorded whether the verdict was confirmed, and assigned unconfirmed findings to a fixed error taxonomy. 
For \texttt{DUPLICATE} findings, the reviewer counted semantically equivalent occurrences in the timeline containing the additional row and used the report to determine whether the extra occurrence was an
extraction error or a true recurrence; exact wording and identical timestamps were not required.
Confirmation rate was defined over reported findings and therefore does not measure discrepancies the judge failed to emit. 
The 10 reports used in Stage 1 were excluded from the Stage 2 sample, so judge validation and the eight-source leaderboard used disjoint reports.

A second reviewer independently assessed a random sample of 150 findings drawn from 5 of the 10 reports, stratified by extractor pairing. The sample comprised 100 findings from GPT-5 versus o4-mini, 23 from Llama 3.3 70B versus GPT-5, and 27 from Llama 3.3 70B versus o4-mini. Findings were sampled randomly within each pairing. The second reviewer received the case report, timelines, the finding, and the verdict, but was blinded to the primary reviewer's determinations.
Agreement measures, reported per judge, included raw agreement, Cohen's $\kappa$, and Gwet's AC1.

\textbf{Stage 2: eight-source leaderboard.} We conducted a complete round robin over 10 randomly selected reports that did not overlap with Stage 1. The eight entrants were two human annotators (A1 and A2), three proprietary models (GPT-5, Claude Opus 5, and o4-mini), and three open-weight models (GLM-5.2, Gemma 4, and Llama 3.3 70B). All 28 unordered pairings were attempted for each report, giving 280 judge calls. Of these, 279 returned schema-conforming output, and 278 contained at least one decisive finding and entered the fit; the remaining call returned malformed JSON and the other returned valid JSON but contained no \texttt{A} or
\texttt{B} verdict.

\textbf{Stage 3: corpus-scale best-in-class comparison.} The highest-rated human annotator, proprietary model, and open-weight model from Stage 2 were carried forward to all 126 usable reports. Their three pairings were attempted for every report, giving 378 judge calls, of which 376 returned schema-conforming output and 372 contained at least one decisive finding. This stage tested whether the separation and error profiles observed in the 10-report round robin persisted at corpus scale.

\textbf{Stage 4: adjudication-guided merging.} The highest-rated Stage 2 source served as the base timeline. Findings in which either of the other two class-leading sources was supported over the base were converted into edit suggestions containing the disputed base event, the source-supported correction, and the supporting report sentence. A GPT-5.6-sol merger received the report, the base timeline, and the suggestions, and returned an updated event--time sequence with an applied-or-skipped change log. We generated merged timelines for all 126 reports and compared each with its original GPT-5 base using DeepSeek V3.2 and the GAVEL protocol.
For each suggestion, the change log recorded whether the merger skipped the proposal or applied it by adding an event, removing an unsupported or duplicate event, revising a timestamp, or revising a clinical value.

\begin{figure}[t] 
\centering 
\includegraphics[width=0.8\linewidth]{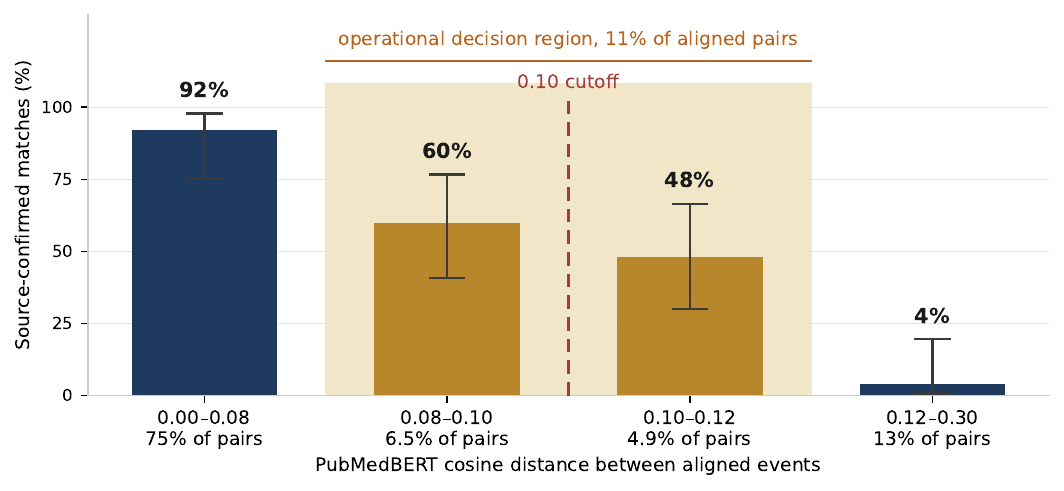} 
\vspace{-2mm}
\caption{Source-confirmed match rates across PubMedBERT cosine-distance bands among 100 aligned event pairs (25 per stratum). 
Bars show the percent judged to describe the same clinical event; whiskers denote Wilson 95\% confidence intervals. The shaded 0.08--0.12 region spans the 0.10 cutoff used in prior timeline evaluations and contains approximately 11\% of all aligned pairs.}
\label{fig:matcher_diagnostic}
\vspace{-2mm}
\end{figure}
\vspace{-2mm}
\subsection*{Scoring and Statistical Analysis}
\vspace{-2mm}
Verdicts \texttt{A} and \texttt{B} contributed weighted points to the supported candidate; \texttt{BOTH}, \texttt{NEITHER}, and \texttt{UNCLEAR} favored neither side. Weights were 3.0 for unsupported additions; 2.0 for wrong times, wrong values, and false duplicates; 1.0 for omitted supported positive findings; and 0.5 for omitted pertinent negatives and recurrences. 
A candidate's point share was its weighted total divided by the combined points assigned to A and B in that game. Games with equal nonzero totals received a point share of 0.5. Judge calls with no \texttt{A} or \texttt{B} verdict were excluded from Bradley--Terry
fitting.
We repeated the Stage 2 analysis with timing-weighted and event-weighted alternatives to test whether the rank order depended on the primary
discrepancy weights.

We fitted Bradley--Terry models to game-level point shares \cite{bradley1952rank}. Fitted abilities were converted to ratings as $R_i=1500+(400/\ln 10)(\beta_i-\bar{\beta})$, where $\beta_i$ is the fitted log-ability of entrant $i$ and $\bar{\beta}$ is the mean fitted ability across entrants. Each leaderboard is therefore centered at 1500, and only differences between entrants within the same analysis are interpretable. Confidence intervals were obtained from 2,000 report-level bootstrap resamples, preserving all within-report pairings. A finding was assigned to an entrant when the source-supported adjudication contradicted that entrant's account and was categorized as wrong time, wrong value, missed positive event, missed negative finding, missed recurrence, false duplicate, or over-annotated event. These are pairwise discrepancy instances rather than deduplicated errors, so the same underlying event may appear against multiple opponents. Stage 2 rates were summed across an entrant's seven opponents and divided by the 10 reports; Stage 3 rates were normalized per entrant--report matchup; and Stage 4 rates per report. \texttt{UNCLEAR} findings were excluded from rankings and error summaries. 


\begin{table}[t] 
\centering 
\vspace{2mm}
\caption{Manual validation and error taxonomy for GAVEL judge findings. Confirmation rate is calculated among findings reviewed by the primary reviewer. The five categories grouped as ``Other'' were malformed finding, false one-sided finding, wrong placement credited, matching error, and indeterminate timing.} 
\vspace{-5pt}
\label{tab:judge_validation} 
\footnotesize
\setlength{\tabcolsep}{5pt}
\begin{tabular}{lrrr} 
\toprule 
\textbf{Error type} & \textbf{GPT-5.6-sol} & \textbf{DeepSeek V3.2} & \textbf{Total} \\ 
\midrule 
Timing ruled unclear & 59 & 67 & 126 \\ 
Undercalled over-extraction & 64 & 9 & 73 \\ 
Anchor forced on two-sided case & 21 & 14 & 35 \\ 
Missed one-sided finding & 1 & 13 & 14 \\ 
Overcalled duplicate & 7 & 7 & 14 \\ 
Perception timing & 6 & 8 & 14 \\ 
Other (5 categories) & 2 & 13 & 15 \\ 
Value miscalled & 4 & 6 & 10 \\ 
\midrule 
\textbf{Total errors} & \textbf{164} & \textbf{137} & \textbf{301} \\ 
\textbf{Findings reviewed} & \textbf{1,540} & \textbf{1,198} & \textbf{2,738} \\
\textbf{Confirmation rate} & \textbf{89.4\%} & \textbf{88.6\%} & \textbf{89.0\%} \\
\bottomrule 
\end{tabular} 
\vspace{-3mm}
\end{table}

\section*{Results}
\vspace{-3mm}

\textbf{Matcher Discrimination Deteriorated Near the Operational Threshold.}
PubMedBERT cosine distance separated clear matches from mismatches at the extremes but discriminated poorly around the 0.10 threshold (Figure~\ref{fig:matcher_diagnostic}). Of 25 sampled pairs below 0.08, 23 were true matches (92.0\%; 95\% CI, 75.0--97.8\%), compared with 1 of 25 at 0.12--0.30 (4.0\%; 95\% CI, 0.7--19.5\%). Immediately around the cutoff, 15 of 25 pairs at 0.08--0.10 (60.0\%; 95\% CI, 40.7--76.6\%) and 12 of 25 at 0.10--0.12 (48.0\%; 95\% CI, 30.0--66.5\%) were true matches. The middle-band intervals substantially overlapped and together contained approximately 11\% of the 40,259 aligned pairs.


\textbf{Findings from Two GAVEL Judge Models Had Similar Confirmation Rates.}
Across the same 30 validation comparisons adjudicated independently by both judge models, the primary reviewer assessed 2,738 emitted findings (Table~\ref{tab:judge_validation}). GPT-5.6-sol produced 1,540 findings, of which 1,376 were confirmed (89.4\%); DeepSeek V3.2 produced 1,198, of which 1,061 were confirmed (88.6\%). The two judge models had similar confirmation rates, although DeepSeek
V3.2 generated 342 fewer findings. The confirmation rates describe correctness among reported findings; discrepancy coverage was not measured.
Schema conformance was 99.6\% in Stage 2 (279/280 calls) and 99.5\% at corpus scale (376/378); failures resulted from invalid control characters introduced by tabular text inside evidence strings. Among 14,213 findings with quoted evidence, 99.7\% were locatable after whitespace and quotation-mark normalization, while 89.7\% matched byte for byte. The 37 unlocated spans represented 12 distinct quotations, 9 of which reconstructed laboratory tables by prefixing column headers to data rows.

Of the 301 errors, 126 (41.9\%) were timing disagreements incorrectly assigned \texttt{UNCLEAR}, followed by 73 (24.3\%) undercalled over-extractions and 35 (11.6\%) favoring one anchor interpretation in genuinely two-sided cases;
together these accounted for 77.7\% of errors. Undercalled over-extraction was concentrated in GPT-5.6-sol (64 vs.\ 9), whereas DeepSeek V3.2 more often missed one-sided findings (13 vs.\ 1).
Independent reviewer agreement was 89.8\% for GPT-5.6-sol ($\kappa=0.59$, AC1$=0.87$) and 96.4\% for DeepSeek V3.2 ($\kappa=0.78$, AC1$=0.96$). Agreement was 91.0\% for GPT-5 versus o4-mini and 95.6\% for pairings involving Llama 3.3 70B, consistent with finer-grained disagreements being harder to adjudicate. The primary reviewer judged 21 of 128 GPT-5.6-sol findings and 7 of 84 DeepSeek V3.2 findings to be incorrect; the second reviewer judged 16 and 8 to be incorrect, respectively. Remaining reviewer disagreements were concentrated in contested anchors and duplicate calls. 

\begin{figure}[t]
\centering
\includegraphics[width=0.9\linewidth]{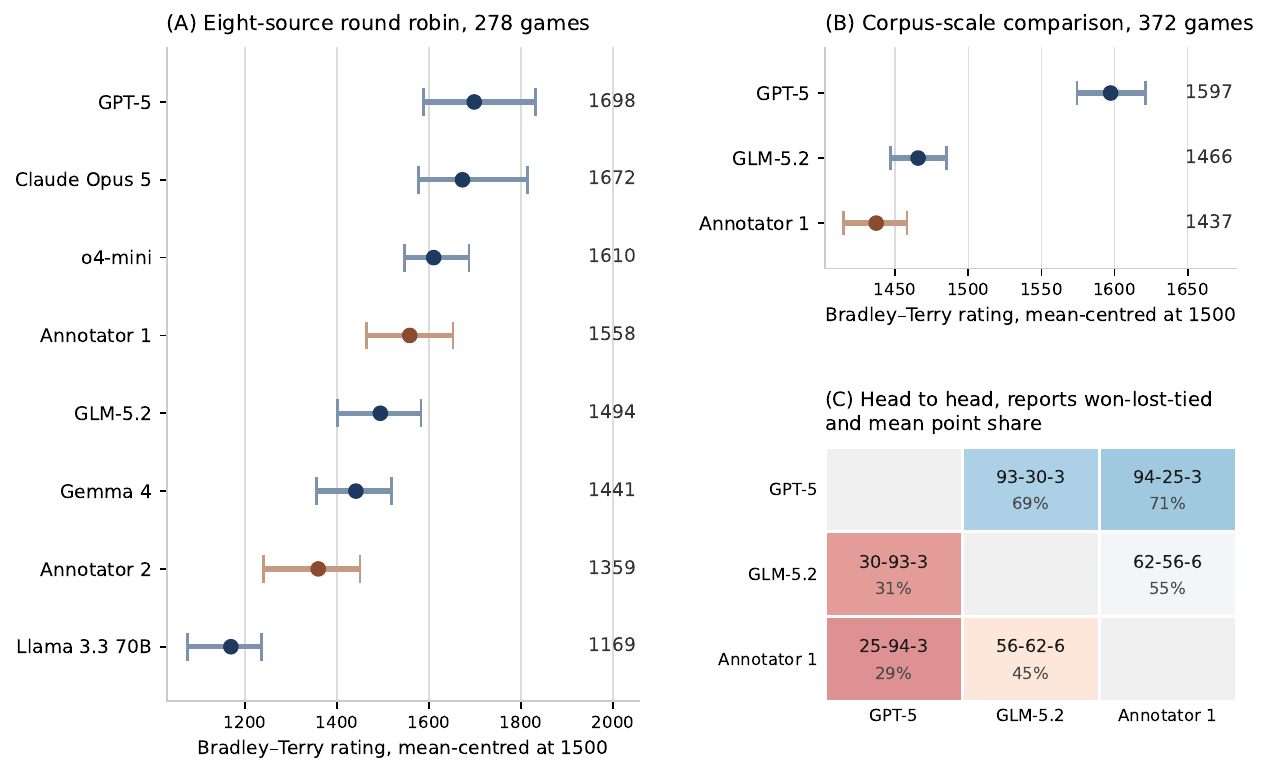}
\caption{Bradley--Terry ratings with case-bootstrap 95\% intervals:(A) eight-source round robin; (B) corpus-scale class-leader comparison; and (C) head-to-head record among the corpus-scale entrants, giving reports won, lost, and tied with mean share of weighted discrepancy points.}
\vspace{-1.5mm}
\label{fig:leaderboards}
\end{figure}

\textbf{Eight-Source Leaderboard and Error Profiles}
Across the 10-report Stage 2 round robin, GPT-5 received the highest Bradley--Terry rating (1698), followed by Claude Opus 5 (1672) and o4-mini (1610); their adjacent bootstrap intervals overlapped, supporting a leading tier rather than a strict ordering (Figure~\ref{fig:leaderboards}A). Annotator 1 ranked fourth (1558), ahead of all three open-weight models, whereas Annotator 2 ranked seventh (1359), ahead of only Llama 3.3 70B (1169). The annotators' intervals did not overlap (A1, 1463--1652; A2, 1240--1450); among adjacent entrants, only A2 and Llama 3.3 70B had nonoverlapping intervals.

Error profiles differed despite similar aggregate ratings (Table~\ref{tab:error_profiles}A). Missed positive events (mean, 6.1 per case) and wrong times (mean, 4.8) dominated, whereas unsupported additions were uncommon (maximum, 0.2). Claude Opus 5 and o4-mini minimized timing errors but missed more supported events; GPT-5 achieved the lowest total error rate (6.0) by balancing event recovery and temporal placement. Llama 3.3 70B had the highest rates of timing and omission errors, while A2 had the highest wrong-value rate.

\textbf{Corpus-Scale Comparison of GPT-5, GLM-5.2, and Annotator 1}
Across 126 reports, Stage 3 produced 14,493 findings (Figure~\ref{fig:leaderboards}B). GPT-5 with a rating of 1597 had separate bootstrap intervalsfrom GLM-5.2 (1466) and Annotator 1 (1437). Of all findings, 9,910 favored one candidate and entered the error analysis; 4,583 (31.6\%) did not favor either candidate, and 16 games were tied.

GLM-5.2 and A1 showed complementary error profiles (Table~\ref{tab:error_profiles}B): GLM-5.2 had fewer wrong-time findings (2.6 vs.\ 5.5 per entrant--report matchup) but more than twice as many missed positive events (11.5 vs.\ 5.5), while A1 had the highest false-duplicate rate (1.1). GPT-5 had the lowest total error rate (8.1 vs.\ 16.7 for GLM-5.2 and 14.6 for A1) and the lowest or joint-lowest rate in four of seven categories. Unsupported additions remained uncommon (0.0--0.3 per matchup). Figure~\ref{fig:leaderboards}C shows the underlying head-to-head records; GLM-5.2 and A1 split reports nearly evenly, although GLM-5.2 had the larger mean point share.
Sensitivity analysis produced the same rank order under all three discrepancy-weighting schemes (Table~\ref{tab:rubric_sensitivity}); the largest displayed rating shift was 34 points, for o4-mini under the event-weighted rubric, and remained well within the bootstrap intervals.

\begin{table}[t]
\centering
\footnotesize
\caption{Leaderboard stability across three scoring rubrics. The default rubric weights over-annotation 3.0, false duplicates, wrong times and wrong values 2.0, missed positives 1.0, and missed negatives and recurrences 0.5. The timing rubric raises wrong time to 6.0 and lowers the omission categories to 0.2. The events rubric raises over-annotation and false duplicates to 6.0 and all omission categories to 3.0 while lowering wrong time to 1.5. Ratings are Bradley--Terry abilities on the mean-centred scale with case-bootstrap 95\% intervals.}
\vspace{-1.5mm}
\label{tab:rubric_sensitivity}
\footnotesize
\begin{tabular}{lccc}
\toprule
\textbf{Entrant} & \textbf{Default} & \textbf{Timing-weighted} & \textbf{Event-weighted} \\
\midrule
GPT-5 & 1698 (1588--1831) & 1692 (1544--1868) & 1701 (1617--1809) \\
Claude Opus 5 & 1672 (1577--1813) & 1678 (1564--1852) & 1668 (1585--1785) \\
o4-mini & 1610 (1546--1687) & 1642 (1559--1743) & 1576 (1515--1652) \\
Annotator 1 & 1558 (1463--1652) & 1560 (1444--1667) & 1558 (1466--1651) \\
GLM-5.2 & 1494 (1402--1582) & 1509 (1408--1592) & 1492 (1415--1577) \\
Gemma 4 & 1441 (1354--1519) & 1438 (1342--1517) & 1428 (1344--1509) \\
Annotator 2 & 1359 (1240--1450) & 1343 (1222--1426) & 1389 (1271--1482) \\
Llama 3.3 70B & 1169 (1075--1236) & 1139 (991--1222) & 1187 (1107--1244) \\
\bottomrule
\end{tabular}
\vspace{-2mm}
\end{table}

\begin{figure}[!htbp]
\centering
\includegraphics[width=0.9\linewidth]{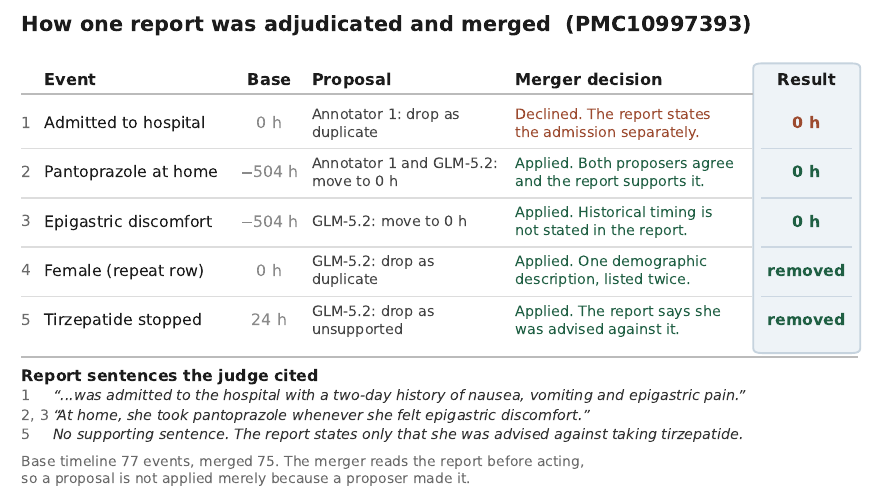}
\vspace{-2mm}
\caption{One report adjudicated and merged. Each row is a proposal made against the GPT-5 base timeline by one of the two proposer timelines, the merger's decision, and the resulting entry. The merger consults the report before acting: the first proposal is declined because the report states the admission separately, while the last is applied because the report supports only that tirzepatide was advised against, not that it was stopped.}
\label{fig:worked_example}
\end{figure}

\textbf{Adjudication-Guided Merged Timelines Were Preferred to GPT-5.}
Across 126 reports, GAVEL generated 2,038 suggested edits to the GPT-5 base timeline; the merger applied 1,800 (88.3\%) and declined 238. Eight reports received no suggestions, the median was 6, and the maximum was 154 — all of which the merger applied. Of all suggestions, 1,255 originated from Annotator 1 and 783 from GLM-5.2; 92 reports received suggestions from both sources, 26 from one, and eight from neither. Figure~\ref{fig:worked_example} shows one report end to end.

When adjudicated by DeepSeek V3.2, the merged timeline was preferred in
97 of 126 report-level comparisons (77.0\%; 95\% CI, 69.8--84.1\%),
GPT-5 was preferred in 11 (8.7\%), and 18 were tied (14.3\%)
(Table~\ref{tab:error_profiles}C). The mean number of
discrepancies decreased from 7.63 per report
(95\% CI, 5.39--10.84) for GPT-5 to 0.85
(95\% CI, 0.57--1.20) for the merged timeline.
Among the 1,069 decisive findings, 962 favored the merged timeline
(90.0\%) and 107 favored GPT-5. The largest reductions were in missed
events (5.35 to 0.35 per report) and wrong times
(1.66 to 0.19). False duplicates increased slightly from 0.10 to 0.13
per report, and missed recurrences increased from 0.00 to 0.03.

\begin{table}[t]
\footnotesize
\centering
\caption{Mean pairwise discrepancy instances by entrant:
(A) Stage 2 per report across seven opponents;
(B) Stage 3 per entrant--report matchup; and
(C) GPT-5 base versus merged timelines per report across 126 reports.
Panel C values are shown to two decimal places. Because these are pairwise
instances, the same underlying error may appear in comparisons with multiple
opponents.}
\vspace{-5pt}
\label{tab:error_profiles}
\footnotesize
\setlength{\tabcolsep}{2.25pt}
\begin{tabular}{@{}lrrrrrrrr@{}}
\toprule
\textbf{Entrant} & \shortstack{\textbf{Wrong}\\\textbf{time}} & \shortstack{\textbf{Missed}\\\textbf{positive}} & \shortstack{\textbf{Missed}\\\textbf{negative}} & \shortstack{\textbf{Missed}\\\textbf{recurrence}} & \shortstack{\textbf{Wrong}\\\textbf{value}} & \shortstack{\textbf{False}\\\textbf{duplicate}} & \shortstack{\textbf{Over-}\\\textbf{annotated}} & \textbf{Total} \\
\midrule
\multicolumn{9}{@{}l}{\textit{A. Eight-source round robin}} \\
GPT-5             & 2.6  & 2.7  & 0.2 & 0.0 & 0.0 & 0.3 & 0.2 & 6.0  \\
Claude Opus 5     & 1.4  & 5.1  & 0.2 & 0.1 & 0.1 & 0.3 & 0.0 & 7.2  \\
o4-mini           & 0.9  & 5.0  & 0.7 & 0.0 & 0.1 & 0.3 & 0.1 & 7.1  \\
Annotator 1 (A1)  & 3.3  & 2.6  & 0.6 & 0.0 & 0.5 & 0.3 & 0.1 & 7.4  \\
GLM-5.2           & 3.2  & 5.7  & 0.4 & 0.1 & 0.7 & 0.7 & 0.1 & 10.9 \\
Gemma 4           & 4.1  & 7.5  & 1.2 & 0.1 & 0.9 & 0.1 & 0.1 & 14.0 \\
Annotator 2 (A2)  & 5.2  & 8.1  & 0.6 & 0.0 & 1.5 & 0.5 & 0.1 & 16.0 \\
Llama 3.3 70B     & 17.4 & 11.7 & 3.1 & 0.7 & 1.3 & 0.3 & 0.1 & 34.6 \\
\cmidrule(lr){1-9}
\emph{Mean entrant}
& \emph{4.8}
& \emph{6.1}
& \emph{0.9}
& \emph{0.1}
& \emph{0.6}
& \emph{0.3}
& \emph{0.1}
& \emph{12.9} \\
\bottomrule
\addlinespace[3pt]
\multicolumn{9}{@{}l}{\textit{B. Corpus-scale comparison}} \\
GPT-5             & 3.7 & 3.2  & 0.4 & 0.1 & 0.1 & 0.5 & 0.2 & 8.1  \\
GLM-5.2           & 2.6 & 11.5 & 1.4 & 0.3 & 0.5 & 0.3 & 0.0 & 16.7 \\
Annotator 1 (A1)  & 5.5 & 5.5  & 1.2 & 0.1 & 0.8 & 1.1 & 0.3 & 14.6 \\
\cmidrule(lr){1-9}
\emph{Mean entrant}
& \emph{3.9}
& \emph{6.8}
& \emph{1.0}
& \emph{0.1}
& \emph{0.5}
& \emph{0.6}
& \emph{0.2}
& \emph{13.1} \\
\bottomrule
\addlinespace[3pt]
\multicolumn{9}{@{}l}{\textit{C. Base versus merged timelines}} \\
GPT-5 base        & 1.66 & 5.35 & 0.21 & 0.00 & 0.06 & 0.10 & 0.25 & 7.63 \\
Merged            & 0.19 & 0.35 & 0.01 & 0.03 & 0.00 & 0.13 & 0.14 & 0.85 \\
\bottomrule
\end{tabular}
\vspace{-10pt}
\end{table}

\section*{Discussion and Conclusion}
\vspace{-3mm}

Our results show that both event alignment and reference selection can affect timeline evaluation. GAVEL compares candidate timelines directly with the case report, so neither timeline must be selected as the reference. Our audit analyses showed that errors in event alignment directly affect event-recovery and
temporal-agreement scores.
Each GAVEL finding links its verdict to the report text used to make it, so reviewers can trace the pairwise score to the evidence for individual discrepancies. The resulting rankings may change under different discrepancy definitions, timing tolerances, or scoring weights.
Treating A1 and A2 as leaderboard entrants allowed their independently constructed timelines to be compared directly with the case reports. Under reference-based scoring, disagreements between the annotators change an extractor's score depending on which annotation is selected as the reference. A1 and A2 had nonoverlapping Bradley--Terry rating intervals, showing that the two annotations are not interchangeable reference standards on this corpus. 

The error categories distinguished systems with similar overall ratings: GLM-5.2 and A1 reached similar overall performance through different error profiles. GLM-5.2 placed events more accurately in time but omitted
more supported events, whereas A1 recovered more events but made more timing and duplicate errors. GPT-5 performed well by balancing event recovery and temporal placement. These profiles indicate whether system development should
target event coverage, timestamp assignment, value extraction, or duplicate handling. In this corpus, omissions and timing errors were the main opportunities for improvement. 

Applying the same protocol with two judge models produced similar confirmation rates among the findings they emitted. Their output volumes and error distributions differed, indicating that judge choice can affect which
discrepancies are identified. The frequent assignment of \texttt{UNCLEAR} to timing differences that the primary reviewer considered resolvable identifies a specific target for refining the adjudication rules and for focused clinical review.

We also used GAVEL findings as edit instructions for the timeline merge experiment. The findings served as targeted edit proposals, and the largest improvements occurred in event coverage and temporal placement, consistent with the error-profile analysis. 
The merger selectively incorporated report-supported events and timestamps from Annotator 1 and GLM-5.2 into the GPT-5 timeline. DeepSeek V3.2 performed the final comparison, so the merge result remains an LLM-based assessment rather than an independent clinical review.

\textbf{Limitations.} Our study has some limitations to consider. \textbf{First}, our study is limited to published case reports from a single drug class GLP1-RA, and these case reports may themselves omit or wrongly state clinical events. 
\textbf{Second}, the matcher audit sampled 25 pairs within each distance band, so it characterizes those bands with limited precision and does not estimate the corpus-wide mismatch prevalence.
\textbf{Third}, validation assessed findings reported by the judge models and therefore estimates correctness among reported findings; discrepancy recall remains unmeasured.
\textbf{Fourth}, the full leaderboard included only 10 case reports, and close rankings may depend on the judge model, candidate order, temporal tolerance, and treatment of non-decisive verdicts.
\textbf{Finally,} pairwise adjudication cannot detect an event omitted or misrepresented identically by both timelines, and merging may preserve these
shared errors. 

\vspace{-2mm}
\subparagraph{Acknowledgments}
This research was supported by the Intramural Research Program of the National Institutes of Health and utilized the computational resources of the \href{http://hpc.nih.gov}{HPC Biowulf cluster}. The contributions of NIH author(s) are considered Works of the United States Government. 
The findings and conclusions in this paper are those of the author(s) and do not necessarily reflect the views of the NIH or the U.S. Department of Health and Human Services. 

\bibliographystyle{vancouver}
\bibliography{references}  

\end{document}